\documentclass{article}
\usepackage{spconf,amsmath,graphicx}

\usepackage{booktabs}
\usepackage{multirow}  
\usepackage{algorithm}
\usepackage{algorithmicx}
\usepackage{algpseudocode}
\usepackage{amsmath}
\usepackage{enumitem}
\usepackage{url}
\setlist{nosep, leftmargin=14pt}

\usepackage{mwe} 

\title{MORPHFED: Federated Learning for Cross-institutional Blood Morphology Analysis}
\name{Gabriel Ansah \qquad Eden Ruffell \qquad Delmiro Fernandez-Reyes \qquad Petru Manescu}
 \address{UCL Department of Computer Science}
\begin{document}
%
\maketitle
\begin{abstract}

Automated blood morphology analysis can support hematological diagnostics in low- and middle-income countries (LMICs) but remains sensitive to dataset shifts from staining variability, imaging differences, and rare morphologies. Building centralized datasets to capture this diversity is often infeasible due to privacy regulations and data-sharing restrictions. We introduce a federated learning framework for white blood cell morphology analysis that enables collaborative training across institutions without exchanging training data. Using blood films from multiple clinical sites, our federated models learn robust, domain-invariant representations while preserving complete data privacy.
Evaluations across convolutional and transformer-based architectures show that federated training achieves strong cross-site performance and improved generalization to unseen institutions compared to centralized training. These findings highlight federated learning as a practical and privacy-preserving approach for developing equitable, scalable, and generalizable medical imaging AI in resource-limited healthcare environments.

\end{abstract}
\begin{keywords}
Federated learning, Vision models, Federated aggregation, Centralized training, Blood Cell morphology analysis, Data privacy 
\end{keywords}
\section{Introduction}
\label{sec:intro}

Microscopic examination of Peripheral Blood Smears (PBS) and Bone Marrow Aspirates (BMA) remains the gold standard for diagnosing and subtyping leukemias, anemias, infections, and inherited blood disorders—particularly where access to advanced molecular diagnostics is limited \cite{yadav_availability_2021}. However, this process is labor-intensive and depends on a shrinking pool of skilled experts, underscoring the need for accessible, scalable, and cost-effective diagnostic solutions, especially in resource-constrained healthcare systems.
Recent advances in deep learning have demonstrated significant potential to automate morphological analysis in PBS and BMA, aiding in the rapid detection of hematological \cite{matek_highly_2021}. Yet, such models are highly sensitive to domain shifts caused by variations in staining, imaging devices, and rare cell morphologies, leading to reduced generalization across laboratories and populations. Achieving robust performance requires diverse, large-scale datasets capturing this variability. However, assembling such datasets typically demands centralized training pipelines, involving aggregation of large volumes of sensitive medical data and access to high-end computational infrastructure \cite{rieke_future_2020}. These requirements raise serious privacy, regulatory, and logistical challenges, especially in low- and middle-income countries (LMICs), where imaging and annotation resources are limited. Consequently, small-sample-size effects and underrepresentation of diverse populations further exacerbate model bias and reduce generalizability \cite{guan_federated_2024}. Moreover, storing and processing large medical imaging datasets often exceeds the computational capacity available in many LMIC clinical settings \cite{yadav_availability_2021}. Therefore, there is a critical need for privacy-preserving, resource-efficient, and collaborative learning strategies that can facilitate the development of reliable diagnostic AI without centralizing data. Federated Learning (FL) offers a promising paradigm to address these challenges by enabling joint model training across multiple institutions without sharing raw data. FL preserves data privacy while leveraging collective knowledge to improve model robustness and generalization. Despite its growing adoption in other medical imaging domains\cite{sun_federated_2025}, its application to blood cell morphology analysis in resource-limited settings remains largely unexplored. Addressing this gap is essential for developing equitable, scalable, and privacy-preserving AI-assisted diagnostic solutions for PBS and BMA analysis.

\section{Methodology}
\label{sec:methodology}

\subsection{Datasets}
\label{subsec:data}
\begin{table}[htb]
\centering
\caption{Class distribution across federated clients.}
\label{tab:class_distribution}
\small
\begin{tabular}{@{}lcccc@{}}
\toprule
\textbf{Cell Type} & \multicolumn{2}{c}{\textbf{Client 1 (JHH)}} & \multicolumn{2}{c}{\textbf{Client 2 (MUH)}} \\
\cmidrule(lr){2-3} \cmidrule(lr){4-5}
& Count & \% & Count & \% \\
\midrule
Band neutrophilis & 164 & 0.8 & 66 & 0.7 \\
Basophil & 42 & 0.2 & 47 & 0.5 \\
Eosinophils & 86 & 0.4 & 254 & 2.8 \\
Lymphocyte & 2,705 & 12.8 & 2,362 & 26.3 \\
Lymphocyte atypical & 350 & 1.7 & 7 & 0.1 \\
Metamyelocyte & 61 & 0.3 & 9 & 0.1 \\
Monocyte & 1,030 & 4.9 & 1,074 & 12.0 \\
Myelocyte & 138 & 0.7 & 25 & 0.3 \\
Promyelocyte & 529 & 2.5 & 42 & 0.5 \\
Segmented neutrophils & 1,911 & 9.0 & 5,090 & 56.7 \\
Smudged cells & 2,267 & 10.7 & 9 & 0.1 \\
\midrule
\textbf{Total} & \textbf{21,200} & \textbf{100.0} & \textbf{8,985} & \textbf{100.0} \\
\bottomrule
\end{tabular}

\end{table}

We used two independent datasets from two different centers with 11 cell types in common (Table~\ref{tab:class_distribution}), ensuring consistent classification targets while maintaining the heterogeneity of the natural distribution essential for the evaluation of federated learning. A third dataset from Hospital Clinic of Barcelona (Client 3, 12,992  images) was held out exclusively for independent external validation, serving to assess model generalization to completely unseen institutional data with distinct imaging protocols and patient populations.
\begin{figure}
    \centering
    \includegraphics[width=0.9\linewidth]{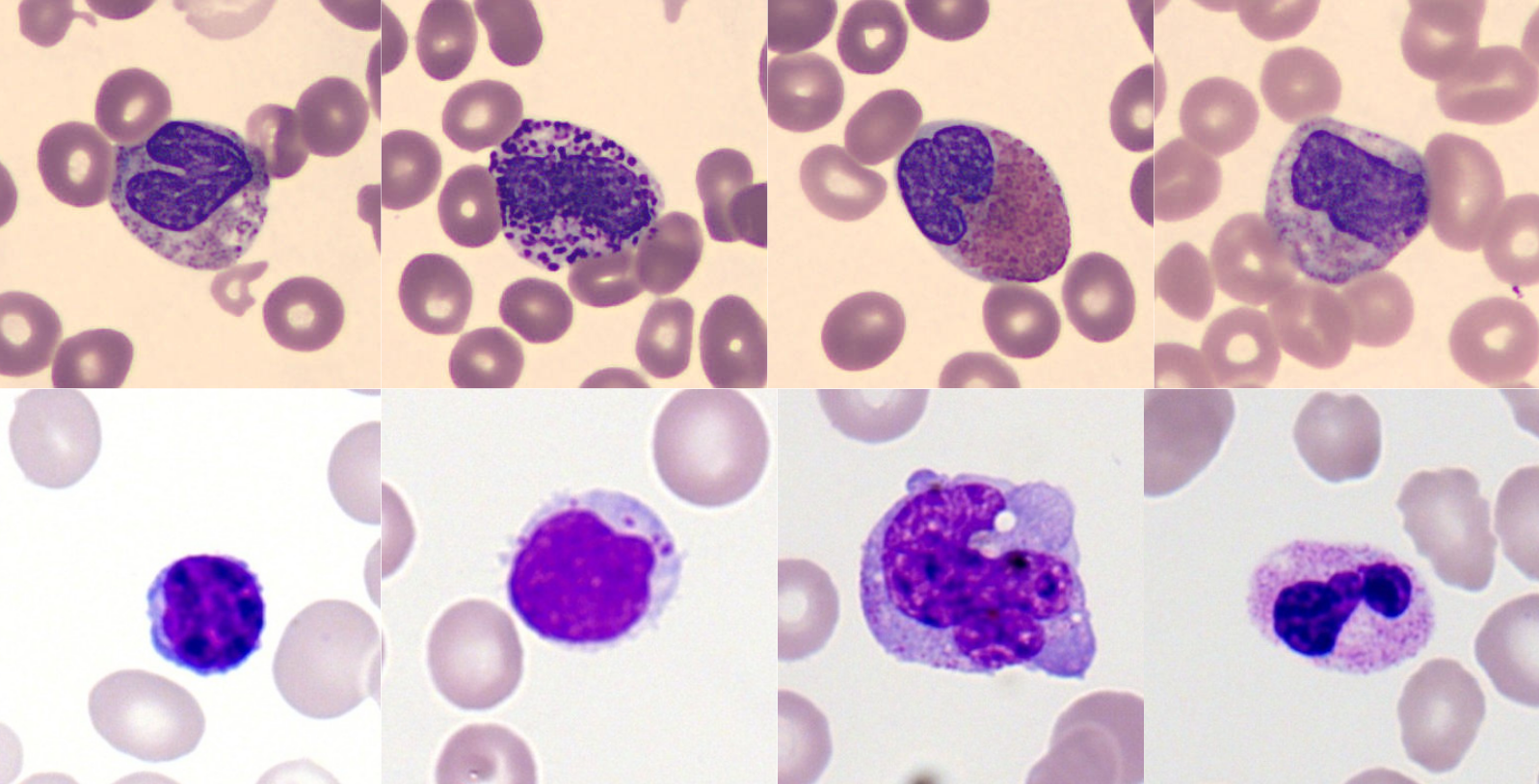}
    \caption{Sample cell types present in the two training datasets. Staining variation can be observed between Client 1 (first row) and Client 2 (second row) datasets }
    \label{fig:placeholder}
\end{figure}

\subsection{Experimental Design}
\label{subsec:experimental}
\begin{figure*}[htb]
    \centering
    \includegraphics[width=\textwidth]{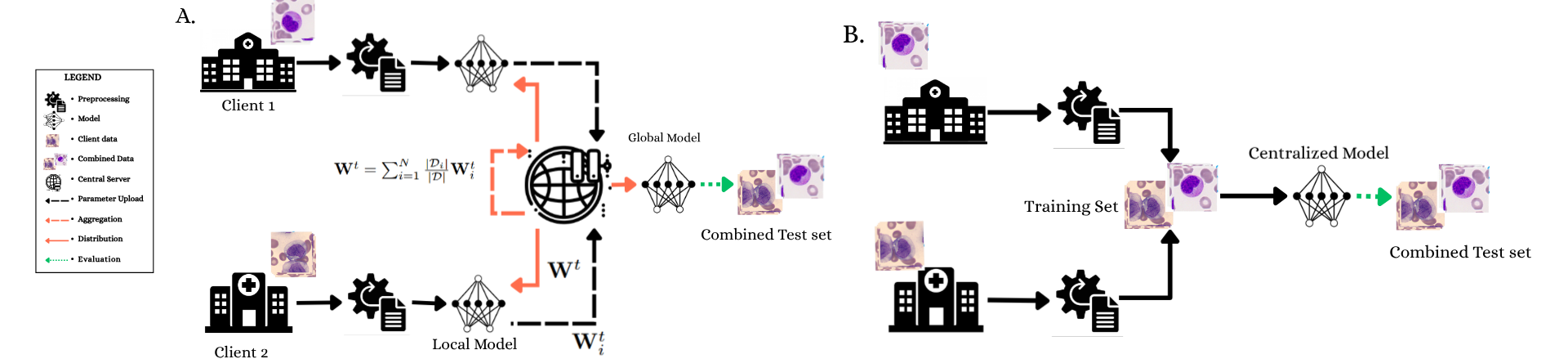}
    \caption{(A) Federated Learning framework demonstrates privacy-preserving collaborative training where Client 1 and Client 2 perform local model training with parameter aggregation at a central server (B) Centralized Training paradigm with full access to combined dataset using 4-fold cross-validation}
    \label{fig:exp_setup}
\end{figure*}
We evaluated three learning paradigms: (1) federated learning across distributed institutions, (2) centralized training with combined data, and (3) Local training with individual client data. Four aggregation strategies are compared: FedAvg, FedMedian, FedProx, and FedOpt. Two architectures are employed: ResNet-34 (CNN baseline with ImageNet pre-training) and DINOv2-Small (self-supervised Vision Transformer). 

Training followed a standardized protocol: federated models used 5 global communication rounds with 5 local epochs per client per round (25 total epochs); centralized baselines use 25 epochs with 4-fold cross-validation. Data is partitioned as 60\% training, 13.33\% validation, 13.33\% local testing, and 13.33\% for global test evaluation. All images were resized to 224$\times$224 pixels with conservative augmentation (random translation $\pm$10\%, rotation $\pm$5$^{\circ}$) to preserve diagnostic morphology. Both architectures used selective fine-tuning: ResNet-34 freezes early layers while training the final three residual blocks ($\sim$11M parameters); DINOv2-Small freezes early transformer blocks (0-7) while training blocks 8-11 ($\sim$9M parameters). Client 3 data remained isolated from all training procedures, serving solely to evaluate the final models' ability to generalize to new institutional sources.

\subsection{Aggregation Strategies}
\label{subsec:aggregation}

Four federated aggregation methods were evaluated for their robustness to data heterogeneity:

\textbf{FedAvg}~\cite{mcmahan_communication-efcient_nodate} computes weighted average of client parameters: $\mathbf{w}_{t+1} = \sum_{i=1}^{N} \frac{n_i}{n}\mathbf{w}_i^t$, where $n_i$ is client $i$'s sample size and $n$ is total samples. This baseline approach is sensitive to outlier updates from clients with extreme class distributions.

\textbf{FedMedian}~\cite{yin_byzantine-robust_2021} applies coordinate-wise median: $\mathbf{w}_{t+1} = \text{median}(\mathbf{w}_1^t, \ldots, \mathbf{w}_N^t)$, providing robustness against Byzantine failures and extreme client heterogeneity by filtering outlier parameters.

\textbf{FedProx}~\cite{li_federated_2020} adds proximal term to local objective: $\min_{\mathbf{w}} F_i(\mathbf{w}) + \frac{\mu}{2}\|\mathbf{w} - \mathbf{w}^t\|^2$, constraining local updates to remain close to global model, improving convergence stability under non-IID data.

\textbf{FedOpt}~\cite{reddi_adaptive_2021} employs adaptive server-side optimization (Adam) on aggregated gradients rather than parameters, dynamically adjusting learning rates to handle heterogeneous client updates and accelerate convergence.

\subsection{Federated Learning Implementation}
\label{subsec:fl_implementation}

We used Flower~\cite{beutel2020flower} with synchronous communication. The central server coordinates training without accessing raw data, distributing global parameters and applying aggregation strategies. Clients train locally and return only parameter updates. To address severe class imbalance (Table~\ref{tab:class_distribution}), we employed Focal Loss~\cite{lin_focal_2018} with modulating factor $(1-p_t)^\gamma$, weighted random sampling, and gradient accumulation over 4 steps (effective batch size 32). Gradient clipping (max norm 1.0) ensures stable convergence.

Performance was evaluated on balanced accuracy, focusing on cross-institutional generalization, assessing robustness when encountering data from institutions with different imaging protocols and patient populations.

\section{Results and Analysis}
\label{sec:results}
\subsection{Evaluation on the Combined Test set}
The initial experiments focused on training the federated learning framework with different aggregation methods to assess which method is best suited for the specific domain tackled in this paper, highly imbalanced and heterogeneous medical data. The models were evaluated on a combined dataset containing data from both clients. 
The results, as presented in Table~\ref{tab:fl_results}, revealed significant architecture-dependent behavior among the aggregation methods. Most notably, FedOpt exhibited extreme variability. It achieved significantly poor performance on ResNet34 (0.3638 balanced accuracy) while maintaining competitive performance on DINOv2-S (0.5594 balanced accuracy). In contrast, FedAvg and FedProx maintained relatively stable performance across both models. FedMedian demonstrated the most consistent performance across both architectures, achieving balanced accuracies of 0.5738 (ResNet34) and 0.5797 (DINOv2-S). 

\begin{table}[htb]
\centering
\caption{Performance comparison of federated learning aggregation methods for ResNet-34 and DINOv2-Small architectures across four federated strategies.}
\label{tab:fl_results}
\begin{tabular}{llcc}
\hline
\textbf{Aggregation} & \textbf{Model} & \textbf{Balanced} & \textbf{Macro} \\
\textbf{Method} & & \textbf{Accuracy} & \textbf{F1-Score} \\
\hline
\multirow{2}{*}{FedAvg} 
    & ResNet-34 & 0.5679 & 0.57 \\
    & DINOv2-S & 0.5591 & 0.47 \\
\hline
\multirow{2}{*}{FedMedian} 
    & ResNet-34 & 0.5738 & 0.56 \\
    & DINOv2-S & \textbf{0.5797} & 0.48 \\
\hline
\multirow{2}{*}{FedProx} 
    & ResNet-34 & 0.5546 & 0.54 \\
    & DINOv2-S & 0.5718 & 0.45 \\
\hline
\multirow{2}{*}{FedOpt} 
    & ResNet-34 & 0.3638 & 0.36 \\
    & DINOv2-S & 0.5594 & \textbf{0.51} \\
\hline
\end{tabular}
\end{table}


The results show that federated learning significantly improved performance compared with models trained only on local institutional data (58\% vs 52\% balanced accuracy), demonstrating the benefit of collaborative training without data sharing. Although federated models perform below a fully centralized model trained on pooled data, they achieve comparable accuracy while preserving complete data privacy.

\begin{table}[htb]
\centering
\caption{Performance comparison on combined test dataset across training paradigms. Federated learning substantially outperforms local training while retaining 87\% (DINOv2-S) and 93\% (ResNet-34) of centralized performance.}
\label{tab:centralized_results}
\small
\begin{tabular}{@{}llcc@{}}
\toprule
\textbf{Model} & \textbf{Training Configuration} & \textbf{Accuracy} & \textbf{Bal. Acc} \\
\midrule
\multirow{4}{*}{DINOv2-S} 
    & Local - Client 1 & 0.6373 & 0.5152 \\
    & Local - Client 2 & 0.7929 & 0.4679 \\
    & Federated (FedMedian) & 0.8628 & 0.5797 \\
    & Centralized (Combined) & \textbf{0.8907} & \textbf{0.6651} \\
\midrule
\multirow{4}{*}{ResNet-34} 
    & Local - Client 1 & 0.6057 & 0.4497 \\
    & Local - Client 2 & 0.5965 & 0.4106 \\
    & Federated (FedMedian) & 0.8415 & 0.5738 \\
    & Centralized (Combined) & \textbf{0.8530} & \textbf{0.6165} \\
\bottomrule
\end{tabular}
\end{table}

However, balanced accuracy metrics do not reveal the complete performance picture regarding class-specific challenges. Table~\ref{tab:global_f1_comparison} presents class-wise F1-scores for the best performing model and aggregation methods, revealing critical insights into minority class performance. Although FedMedian achieves the highest balanced accuracy on DINOv2-S, it completely failed to classify Metamyelocytes (F1: 0.00), a critical diagnostic marker for acute promyelocytic leukemia, and shows poor performance on other minority classes such as Band neutrophils (F1: 0.13). For DINOv2-S, FedOpt emerges as the superior method when considering minority class performance, achieving F1-scores of 0.14 for Metamyelocytes, 0.42 for Basophils, and 0.20 for Band neutrophils, demonstrating better preservation of clinically significant rare cell detection.

\begin{table*}[htb]
\centering
\caption{Class-wise F1-score comparison on Global Test Set (Combined Client 1 and Client 2, 3,477 images) across local, federated, and centralized training paradigms for DINOv2-Small and ResNet-34 architectures.}
\label{tab:global_f1_comparison}
\small
\setlength{\tabcolsep}{3pt}
\begin{tabular*}{\textwidth}{@{\extracolsep{\fill}}lcccccccc@{}}
\toprule
 & \multicolumn{2}{c}{\textbf{DINOv2 Local}} & \multicolumn{2}{c}{\textbf{DINOv2 Federated}} & \textbf{DINOv2} & \textbf{ResNet-34} & \\
\textbf{Cell Type} & \textbf{Client 1} & \textbf{Client 2} & \textbf{FedMed} & \textbf{FedOpt} & \textbf{Central.} & \textbf{Central.} & \textbf{Images} \\
\midrule
Band neutrophilis & 0.13 & 0.19 & 0.13 & 0.20 & \textbf{0.28} & 0.17 & 36 \\
Basophil & 0.18 & 0.25 & 0.35 & 0.42 & \textbf{0.47} & 0.44 & 18 \\
Eosinophils & 0.13 & 0.77 & 0.42 & 0.65 & \textbf{0.87} & 0.73 & 71 \\
Lymphocyte & 0.79 & 0.90 & 0.90 & 0.92 & \textbf{0.93} & 0.91 & 976 \\
Lymphocyte atypical & 0.28 & 0.08 & 0.22 & 0.14 & \textbf{0.44} & 0.48 & 63 \\
Metamyelocyte & 0.11 & 0.00 & 0.00 & 0.14 & \textbf{0.21} & 0.19 & 12 \\
Monocyte & 0.64 & 0.69 & 0.80 & 0.79 & \textbf{0.89} & 0.84 & 410 \\
Myelocyte & 0.19 & 0.31 & 0.35 & 0.31 & \textbf{0.37} & 0.21 & 29 \\
Promyelocyte & 0.54 & 0.49 & \textbf{0.55} & 0.53 & 0.60 & \textbf{0.55} & 97 \\
Segmented neutrophils & 0.80 & 0.91 & 0.82 & 0.92 & \textbf{0.97} & 0.95 & 1384 \\
Smudged cells & 0.47 & 0.68 & \textbf{0.77} & 0.65 & 0.83 & 0.82 & 381 \\
\bottomrule
\end{tabular*}
\end{table*}
Local training consistently performed poorly in all classes compared to both federated approaches, with particularly severe deficiencies in minority classes.
These results quantify the trade-off between privacy preservation and diagnostic accuracy, establishing that federated learning achieves 87\% of centralized performance while providing complete data privacy, representing a viable compromise between institutional data sovereignty and collaborative learning benefits.

\subsection{Evaluation on Out-of-Distribution Data}
\label{sec:ood}
Evaluation on Client 3's external validation dataset from Barcelona (Table~\ref{tab:client3_classwise}) reveals both federated approaches (FedMedian and FedOpt) achieved better generalization on completely unseen institutional data (67\% balanced accuracy) compared to centralized training (64\%). This  suggests that exposure to heterogeneous institutional characteristics during federated training, such as imaging equipment, patient populations, and staining methods~\cite{haller_handling_2023}, may promote learning of more generalizable morphological features. FedMedian demonstrates particularly dramatic improvements on Band neutrophils (F1: 0.62 vs. centralized 0.30, +107\%) and Promyelocytes (0.61 vs. 0.35, +74\%), indicating successful preservation of diagnostically relevant features across varying institutional protocols. However, Metamyelocytes remained challenging for all approaches (F1: 0.02-0.30), reflecting the fundamental difficulty of learning robust representations from extremely rare classes~\cite{sidhom_deep_2021}.
\begin{table}[htb]
\centering
\caption{Class-wise F1-scores on Client 3 external validation (Barcelona, 12,992 images)}
\label{tab:client3_classwise}
\small
\begin{tabular}{@{}lccc@{}}
\toprule
\textbf{Cell Type} & \textbf{FedMed} & \textbf{FedOpt} & \textbf{Centralized} \\
\midrule
Band neutrophilis & \textbf{0.62} & 0.53 & 0.30 \\
Basophil & 0.78 & 0.80 & \textbf{0.85} \\
Eosinophil & 0.90 & \textbf{0.96} & 0.92 \\
Lymphocyte & \textbf{0.86} & 0.78 & \textbf{0.86} \\
Metamyelocyte & 0.02 & 0.11 & \textbf{0.30} \\
Monocyte & 0.82 & \textbf{0.84} & 0.79 \\
Myelocyte & 0.33 & 0.51 & \textbf{0.61} \\
Promyelocyte & \textbf{0.61} & 0.55 & 0.35 \\
Seg. neutrophils & 0.66 & \textbf{0.71} & 0.61 \\
\midrule
\textbf{Accuracy} & 0.72 & \textbf{0.73} & 0.70 \\
\textbf{Bal. Accuracy} & \textbf{0.67} & 0.67 & 0.64 \\
\bottomrule
\end{tabular}
\end{table}

\section{Discussion}
\label{sec:discussion}

This study demonstrates that federated learning can achieve near-centralized performance while fully preserving data privacy, consistent with recent reports in medical imaging~\cite{soltan_scalable_2024}. 
Federated models exhibited better performance when tested on images from unseen institutions, suggesting that distributed training on heterogeneous staining, imaging, and patient distributions promotes faster learning of domain-invariant morphological features. Architecture-aggregation interactions reveal critical design considerations. FedOpt's adaptive optimization amplifies gradient conflicts arising from non-IID data distributions, causing ResNet34's sharp loss landscape to diverge~\cite{zhao_federated_2018,li_federated_2020}, while DINOv2-S's pre-trained transformer backbone demonstrates robustness to non-IID distributions (55.94\%). In contrast, FedMedian provided consistent cross-architecture performance but completely failed in Metamyelocytes, as median-based aggregation suppresses weak signals from the rarest classes. 
We identified critical architecture-aggregation interactions: median-based aggregation ensures robustness but systematically disadvantages rare classes, while FedOpt better preserves rare cell signals at the cost of architectural sensitivity.  Overall, these findings position federated learning as a robust, privacy-preserving, and generalizable framework for hematological image analysis.
\section{Acknowledgments}
No funding was received for conducting this study. The authors have no relevant financial or non-financial interests to disclose.

\section{Compliance with Ethical Standards}
This research did not involve any studies with human participants or animals performed by any of the authors. The study used only publicly available datasets; therefore, ethical approval was not required.

\bibliographystyle{IEEEbib}
\bibliography{refs}

\end{document}